\documentclass[runningheads]{llncs}
\usepackage[T1]{fontenc}
\usepackage{graphicx}
\usepackage{amsmath,amsfonts}
\usepackage{algorithm}
\usepackage{array}
\usepackage[caption=false,font=normalsize,labelfont=sf,textfont=sf]{subfig}
\usepackage{textcomp}
\usepackage{stfloats}
\usepackage{url}
\usepackage{verbatim}
\usepackage{graphicx}
\usepackage{cite}
\usepackage{algpseudocode}
\usepackage{mathtools}
\usepackage{pifont}
\usepackage{tabularx}
\usepackage{booktabs}
\begin{document}

\title{Investigating Permutation-Invariant Discrete Representation Learning for Spatially Aligned Images}

\titlerunning{Investigating Permutation-Invariant Discrete Representations}
%
\author{Jamie Stirling\inst{1} \and
Noura Al-Moubayed\inst{1} \and
Hubert P. H. Shum\inst{1} }

\authorrunning{J. Stirling, N. Al-Moubayed and H. P. H. Shum}
%
\institute{Durham University, United Kingdom}
\maketitle              
\begin{abstract}
Vector quantization approaches (VQ-VAE, VQ-GAN) learn discrete neural representations of images, but these representations are inherently position-dependent: codes are spatially arranged and contextually entangled, requiring autoregressive or diffusion-based priors to model their dependencies at sample time. In this work, we ask whether positional information is necessary for discrete representations of spatially aligned data. We propose the permutation-invariant vector-quantized autoencoder (PI-VQ), in which latent codes are constrained to carry no positional information. We find that this constraint encourages codes to capture global, semantic features, and enables direct interpolation between images without a learned prior. To address the reduced information capacity of permutation-invariant representations, we introduce matching quantization, a vector quantization algorithm based on optimal bipartite matching that increases effective bottleneck capacity by $3.5\times$ relative to naive nearest-neighbour quantization. The compositional structure of the learned codes further enables interpolation-based sampling, allowing synthesis of novel images in a single forward pass. We evaluate PI-VQ on CelebA, CelebA-HQ and FFHQ, obtaining competitive precision, density and coverage metrics for images synthesised with our approach. We discuss the trade-offs inherent to position-free representations, including separability and interpretability of the latent codes, pointing to numerous directions for future work.

\end{abstract}
\section{Introduction}
\begin{table}[t]
    \centering
    \scalebox{0.85}{
      \begin{tabular}{lccc}
        \toprule
        Method & Discrete? &$\mathcal{O}({1})$-time sampling? & Interpolation? \\
        \midrule
        BigGAN \cite{donahue2019large}              &  -  & \ding{52} &  -  \\
        VDVAE \cite{child2020very}                  &  -  &  -  &  -  \\
        DDGAN \cite{xiao2021tackling}               &  -  & \ding{52} &  -  \\
        StyleGAN-XL \cite{sauer2022stylegan}        &  -  & \ding{52} & \ding{52} \\
        LDM-4 \cite{Rombach_2022_CVPR}              &  -  &  -  &  -  \\
        ADM-IP \cite{ning2023input}                 &  -  &  -  &  -  \\
        RDM \cite{teng2023relay}                    &  -  &  -  &  -  \\
        WaveDiff \cite{Phung_2023_CVPR}             &  -  &  -  &  -  \\
        VAE \cite{variational-autoencoder}          &  -  & \ding{52} & \ding{52} \\
        $\beta$-VAE \cite{beta-vae}                 &  -  & \ding{52} & \ding{52} \\
        VQ-VAE \cite{vqvae}                         & \ding{52} &  -  &  -  \\
        VQ-GAN (TT) \cite{taming}                   & \ding{52} &  -  &  -  \\
        VQ-GAN (UT) \cite{unleashing}               & \ding{52} &  -  &  -  \\
        \midrule
        PI-VQ (Ours)                                & \ding{52} & \ding{52} & \ding{52} \\
        \bottomrule
    \end{tabular}}
    \caption{Comparison of the capabilities of our method with the state-of-the-art. ``$\mathcal{O}({1})$-time sampling'' indicates methods which can sample an image with a single forward-pass of a neural network.}
    \label{tab:method-features}
\end{table}

\begin{figure*}[t]
    \centering
    \includegraphics[width=0.95\textwidth]{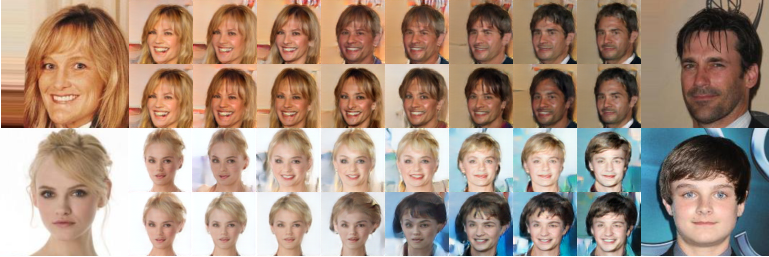}
    \caption{Visualisation of (approximately) smooth interpolations on CelebA 64x64 (source images to left and right). The discrete and permutation-invariant nature of the latents allows us to generate multiple, equally plausible paths between two images.}
    \label{fig:smooth-interpolation}
\end{figure*}

Vector quantization approaches (VQ-VAE, VQ-GAN) have emerged as powerful methods for learning discrete neural representations of images \cite{vqvae,vq-vae-2,taming,vit-vqvae}. However, these learned representations are inherently position-dependent: the same visual feature appearing at different spatial locations may map to different codes, and the emergent ``visual vocabulary'' is highly contextual, requiring autoregressive or diffusion-based priors to model spatial dependencies at sample time \cite{taming,unleashing}. This position- and context-dependence limits the utility of learned codes for interpretability and direct manipulation. It also stands in contrast to the goals of \textit{controllability} and \textit{disentanglement} pursued in the face representation literature \cite{zhao2017dual,bao2018towards,wang2022disentangled}, which aim to identify direct correspondences between interpretable features (such as pose, lighting \cite{deng2020disentangled}, skin tone and subject identity \cite{luo2022styleface}) and factors of variation in learned representations. Such properties have been explored extensively for continuous latent variables in VAEs \cite{beta-vae} and GANs \cite{karras2017progressive,lu2018attribute,gauthier2014conditional}, with recent advances in flow-based approaches \cite{lipman2022flow,dao2023flow} demonstrating efficient sampling. Meanwhile, discrete representations remain comparatively under-explored in this regard. In this work, explore the consequences of enforcing permutation invariance on VQ representations, finding that this constraint encourages codes to capture global, semantic features and enables novel capabilities, including direct interpolation, without any need for learned priors over the latent representations.

In this paper, we propose permutation-invariant discrete representation learning (PI-VQ), a technique for learning position-free discrete representations of spatially aligned face data. We impose a permutation-invariance constraint on the discrete latent codes—a principled inductive bias guaranteeing that no positional information is encoded. We argue that discrete, permutation-invariant representations are especially well-suited to aligned images (such as faces) because: (1) such images exhibit many discrete ground-truth attributes (e.g., eye colour, hair colour, etc. for faces), and (2) spatial alignment reduces the need to encode positional information, allowing codes to focus on global features. As a consequence of this design, we obtain memory-efficient representations that support direct interpolation and $\mathcal{O}(1)
$-time sampling without any need for training a secondary prior over the latent codes.

To complement PI-VQ, we propose \textbf{matching quantization}, a novel vector-quantization algorithm based on optimal bipartite matching. Unlike standard nearest-neighbour quantization \cite{vector-quantization, vqvae}, matching quantization eliminates repeated codes within each image representation, reducing redundancy and increasing the effective information capacity of the permutation-invariant bottleneck.

Finally, we demonstrate that the compositional structure of the learned representations enables \textbf{fast interpolation-based sampling}: novel images can be synthesised by recombining discrete codes from existing images, requiring only a single forward pass. This sampling approach is inherently interpretable—it can be understood as recombining discrete visual \textit{traits}—while approaching competitive scores across multiple quality metrics \cite{fid,precision-recall,density-coverage} (Table \ref{tab:method-features}).

We evaluate PI-VQ on CelebA, CelebA-HQ and FFHQ, showing that the learned representations are concise, semantic, and global. Our method achieves competitive precision, density and coverage metrics, while we discuss limitations on FID and recall in Section~\ref{sec:limitations}.

\noindent Our contributions are as follows:
\begin{itemize}
 \item We propose \textbf{permutation-invariant vector quantization (PI-VQ)}, an architecture for discrete representation learning which enforces that latent codes carry no positional information, and investigate the properties of the resulting representations.
 \item We introduce \textbf{matching quantization}, a vector quantization algorithm based on optimal bipartite matching which eliminates code repetition within each image, multiplying the effective information capacity of the permutation-invariant bottleneck by a factor of 3.5.
 \item We demonstrate that the learned representations enable \textbf{interpolation-based sampling}, allowing us to synthesise novel images in $\mathcal{O}(1)$ forward passes by recombining discrete codes, without requiring a learned prior.
 \item We further show that logistic regression over the learned features is useful for predicting human-annotated features on FFHQ, indicating that the learned representations capture features which separable and interpretable without any explicit disentangling objective.
\end{itemize}

\section{Related Work}

The concept of \textbf{disentanglement} is widespread in deep learning literature \cite{higgins2018towards}, especially in the context of image synthesis \cite{deng2020disentangled} and representation learning \cite{wang2022disentangled}. In continuous representations, disentanglement aims to find low-dimensional vector representations of high-dimensional spaces \cite{caselles2019symmetry} such that the components of the representation correspond to underlying ground-truth factors of variation (such as pose, colour etc. ). VAE \cite{variational-autoencoder} and $\beta$-VAE \cite{beta-vae} have been proposed as robust approaches to disentangled representation learning. Disentanglement is closely related to the idea of \textbf{compositional generalization} \cite{compositional-generalization}, by which a model generalizes to unseen combinations of seen concepts. Recent work has explored disentangling facial motion attributes \cite{tan2024edtalk}, demonstrating the value of separable representations for controllable synthesis.

\textbf{Discrete representation learning} has also emerged as the discrete counterpart to continuous VAE approaches \cite{vqvae}. Discrete representation learning is based on the concept of vector-quantization (VQ) \cite{vector-quantization}, whereby features from a continuous vector space are mapped to an element of a finite set of codebook vectors. This so-called ``VQ-family'' of models includes VQ-VAE \cite{vqvae} and VQ-GAN \cite{taming}. The naive ``nearest-neighbour'' approach to vector quantization usually results in some degree of codebook collapse, in part due to straight-through gradient estimation \cite{vector-quantization}. Recent work has explored improved gradient flow through vector quantization layers \cite{fifty2024restructuring}, addressing limitations of straight-through estimation. Multiple methods have been proposed to mitigate or even eliminate codebook collapse, including EdVAE \cite{baykal2024edvae}, OptVQ \cite{zhang2024preventing}, and HyperVQ \cite{goswami2024hypervq}. Our proposed matching quantization method addresses the problem of codebook collapse in the specific setting of permutation-invariant latent codes. 

\textbf{Permutation-invariant techniques} aim to learn operations which are invariant to the ordering of their inputs \cite{set-transformer}. Empirically, permutation-invariance has been found to have greater robustness to corruption of the inputs and better generalization to unseen situations \cite{permutation-invariant}. Janossy pooling \cite{pi-variable}, Set transformer \cite{set-transformer} and Memory-based Exchangeable Model \cite{pi-mem} have been proposed as specialized architectures for learning operations on sets. Concurrent work has explored permutation-invariant embeddings for automated feature selection in tabular data \cite{liu2025permutation}. While sharing the goal of position-free discrete representations, this approach embeds predefined feature indices for downstream search. To our knowledge, our work is the first to propose a \textit{discrete} analogue to permutation-invariant learning.

\textbf{Sparse binary autoencoders} have been applied in the mechanistic interpretability literature for obtaining sparse, semantic and efficient representations of LLM activations \cite{quirke2025binary}. Our learned discrete set representations of faces can be viewed as a special case of binary autoencoders, with the notable distinction that the fixed code length exactly constrains the sparseness of the representation. 

\section{Methods}
\label{sec:methods}
\begin{figure*}[t]
    \centering
    \includegraphics[width=0.85\linewidth]{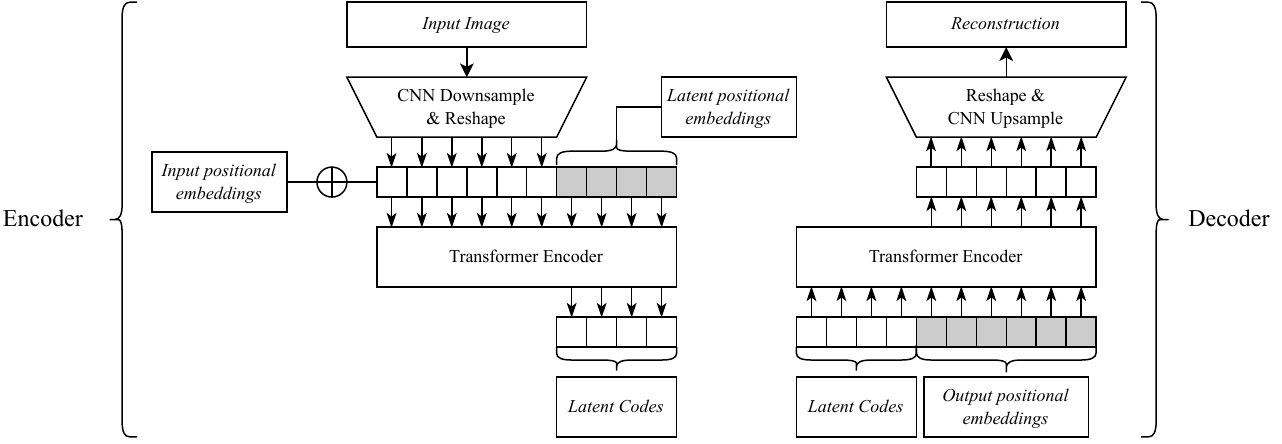}
    \caption{The encoder and decoder components of the permutation-invariant autoencoder (quantization step omitted for clarity). Positional embeddings are not applied to the latent codes before decoding: as a result, the decoder is invariant to permutation of the latent codes.}
    \label{fig:architecture}
\end{figure*}
\subsection{Notation}
The following notation is used from this section onward:
\begin{itemize}
    \item $d$ denotes the dimension of latent embeddings and codebook entries.
    \item $K$ denotes the total number of vector entries in a learned codebook, i.e. codebook $\in\mathbb{R}^{K\times d}$.
    \item $L$ denotes the number of codes used to represent each image (fixed for a given training run).
    \item $x$ and $\hat{x}$ denote an input image and its reconstruction, respectively.
    \item $C_x$ denotes the set of integers that is the discrete coded representation of $x$, such that $|C_x|=L$.
    \item $\begin{psmallmatrix*}
        n \\ r \\
    \end{psmallmatrix*}$ denotes the binomial coefficient used in combinatorics \cite{combinatorics}, defined as $\frac{n!}{r!(n-r)!}$.
\end{itemize}

\subsection{Permutation-Invariant Autoencoder}

The first step in formulating the PI-VQ model is to identify a differentiable sequence of operations which can map the input space to a set of vectors (which lack explicit positional information) and then back to the input space. In addition to the improved generalization and robustness of permutation-invariant representations \cite{permutation-invariant,liu2025permutation}, this architectural constraint forces codes to capture global information about an image. This is particularly appropriate for aligned faces, where many salient features are inherently global (lighting, skin-tone, expression) rather than spatially localized.

To achieve permutation invariance, we exploit the fact that the transformer encoder architecture is invariant to the permutation of its inputs by design \cite{transformer}. We intentionally omit positional embeddings when supplying latent codes to the decoder (Fig.~\ref{fig:architecture}). Enforcing this architectural constraint guarantees that the decoding step is unaffected by the permutation of latent codes.

Below is a description of how the encoder, quantizer and decoder work together to extract discrete permutation-invariant representations of images. Fig.~\ref{fig:architecture} aids in illustrating how information flows through the architecture.

\begin{enumerate}
    \item \textbf{Encoder}: The input image $x$ first goes through a series of convolutional down-sampling and residual layers to produce a $W\times H\times d$ tensor. This is then ``flattened'' into a sequence of $(W\times H)$ vectors of dimension $d$. Input positional embeddings are added, and a further sequence of $L$ learned positional embeddings (also of dimension $d)$ are concatenated. The new sequence is then fed to a transformer encoder, of which the final $L$ output elements are the latent codes.
    \item\textbf{Quantizer}: Each vector in the latent codes is mapped to a similar element of the codebook (Sections \ref{sec:quantization} and \ref{sec:matching}).
    \item\textbf{Decoder}: The quantized latent vectors are concatenated to a sequence of learned output embeddings. The new sequence is fed to a transformer encoder. Of the resulting output sequence, the elements following the first $L$ are reshaped and fed to a series of convolutional up-sampling and residual layers, resulting in the reconstructed image.
\end{enumerate}

\subsection{Vector Quantization}
\label{sec:quantization}
Having identified a suitable architecture for permutation-invariant autoencoder, the next step is to \textit{quantize} the latent representations using a learned mapping $\mathbb{R}^d\rightarrow\mathbb{Z}$. The original VQ-VAE \cite{vqvae} uses nearest-neighbour vector quantization, in which encoder outputs are mapped to their nearest neighbour in a learned codebook. Specifically, for each position-free encoder output vector $z_e$, the corresponding quantized vector is computed as the nearest codebook entry $e_c$, where:
\begin{equation}
    c=\arg\min_j||z_e-e_j||_2
\end{equation}
where $e_0,e_1...e_{K-1}$ are entries in a learned vector codebook of length $K$.

Since the quantization step is non-differentiable, it is necessary to estimate the gradients during backpropagation. For this purpose, we use straight-through gradient estimation \cite{vqvae}, whereby the gradients are copied directly from the decoder input $z_c$ to the encoder output $z_e$.

\subsection{Matching Quantization for Improved Information Bottleneck}
\label{sec:matching}
\begin{figure}
    \centering
    \includegraphics[width=0.65\linewidth]{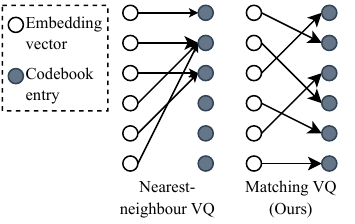}
    \caption{Matching quantization: Our novel approach to vector quantization (right) ensures that no two embedding vectors are mapped to the same codebook element in a given image, effectively minimising redundancy in the discrete representation.}
    \label{fig:matching}
\end{figure}
Naive formulations of VQ-family approaches \cite{vqvae} use the ``nearest-neighbour'' quantization method \cite{vector-quantization} described above. Applying the nearest-neighbour approach alongside PI-VQ, we find in practice that the maximum per-sample codebook usage $K_{\mathrm{img}}$ is much smaller than the number of codes per sample $L$ by the end of training (Table \ref{tab:bottleneck-usage}). This happens because the quantization approach often assigns many embedding vectors to the same codebook entry (Figure \ref{fig:matching}), resulting in redundancy due to repetition. This upper-bounds the per-image information capacity of the latent bottleneck at:
\begin{equation}
    B=\log_2\left[\begin{pmatrix}K_{\mathrm{data}}\\K_{\mathrm{img}}\\ \end{pmatrix}\times \begin{pmatrix}L+K_{\mathrm{img}}-1\\K_{\mathrm{img}}-1\\ \end{pmatrix}\right] \text{bits,}
    \label{eq:nearest-capacity}
\end{equation}
\noindent where $K_{\mathrm{data}}$ is the actual codebook usage over the whole dataset (see supplemental for proof). In practice, nearest-neighbour quantization results in a very low value of $K_{\mathrm{img}}$ (Section \ref{sec:experiments}), so the value of (\ref{eq:nearest-capacity}) is only a fraction of the theoretical maximum (see supplemental for proof) of around 2313 bits ($K=4096$ and $L=512$ ).

To remedy this practical limitation, we propose \textit{matching quantization} as a robust means of improving per-sample codebook usage and the corresponding information capacity. Instead of mapping each latent embedding to its nearest neighbour in the learned codebook, we instead find the minimal-cost \textit{one-to-one matching} between the set of latent embeddings and the elements of the codebook (where cost is the total euclidean distance between latent codes and their matching codebook vector). Formally, this can be expressed as finding the optimal permutation $P\in\{0,1\}^{K\times K}$ of the distance matrix $D\in \mathbb{R}^{K\times L}$ where $D_{ij}$ is the euclidean distance between the codebook entry $i$ and latent embedding $j$ \cite{hungarian}:
\begin{equation}
    P^*=\arg\min_P \mathrm{Tr}(PD) .
\end{equation}

The row index $i$ of the $1$ in the $j^{th}$ column of $P$ is then the index of the codebook entry to which latent embedding $j$ is mapped. In practice, we use the Hungarian algorithm \cite{hungarian} to compute the optimal matching, which is an $\mathcal{O}(n^3)$ algorithm where $n=\max(L,K)$ (compare with the nearest-neighbour quantization which is $\mathcal{O}(n^2)$). We observe that the added complexity of matching quantization (during both training and inference) is a necessary trade-off for the increased effective information capacity.

This approach to vector quantization ensures that the per-image codebook usage is always exactly the size of the representation $L$ for as long as $K\geq L$, i.e., each distinct code is used no more than once per image representation. The corresponding theoretical information capacity is then:
\begin{equation}
    B=\log_2\left[\begin{pmatrix}K_{\mathrm{data}}\\L\\ \end{pmatrix}\right] \text{ bits,}
    \label{eq:matching-capacity}
\end{equation}
where $K_{\mathrm{data}}$ is the actual codebook usage over the dataset observed at the end of training (typically smaller than the hyperparameter $K$ \cite{vq-robust}). This theoretical upper bound on information capacity is derived from the fact that there are $K_{\mathrm{data}}$ codebook entries from which to choose when representing an image, of which $L$ are chosen to represent a single image, resulting in a theoretical representation space of cardinality $\begin{psmallmatrix*}K_{\mathrm{data}}\\L\\ \end{psmallmatrix*}$.

\begin{figure}
    \centering
    \includegraphics[width=0.75\linewidth]{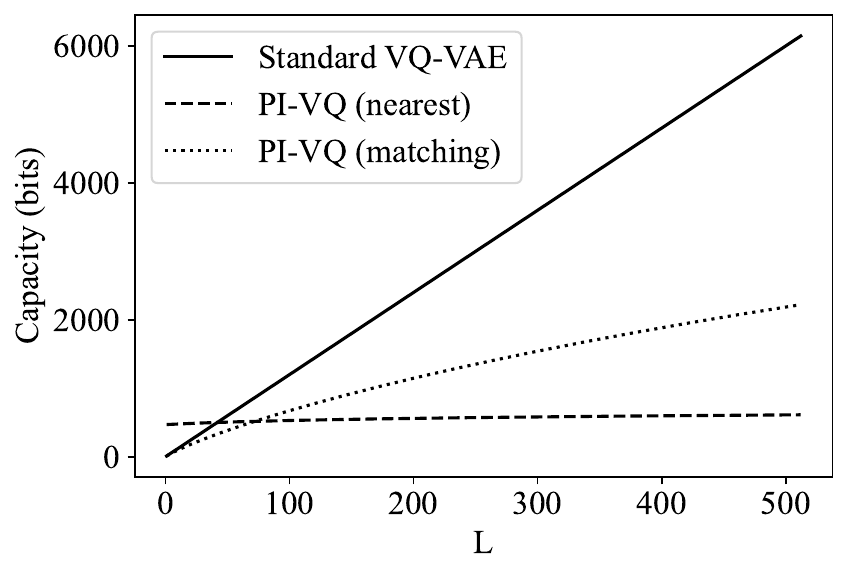}
    \caption{Information capacity (in bits) in terms of representation length $L$ of three approaches with codebook size $K=4096$: standard VQ-VAE with no permutation invariance; PI-VQ with nearest neighbour quantization ($K_{\mathrm{img}}=49$); PI-VQ with proposed matching quantization.}
    \label{fig:capacity}
\end{figure}

We compare information capacities from different settings by plotting the upper bound for both quantization methods on the CelebA-HQ dataset, with the standard VQ-VAE capacity included for reference \ref{fig:capacity}. Figure \ref{fig:capacity} also indicates that PI-VQ information capacity is strictly smaller than standard VQ-family approaches. We address this practically by setting a higher $L$ and $K$ in our experiments (Section \ref{sec:experiments}) than is used in the literature \cite{taming} for a comparable VQ-VAE or VQ-GAN training run, resulting in an overall similar information capacity.

\subsection{Fast Interpolation-Based Sampling}

Besides increasing the latent bottleneck's information capacity, matching quantization facilitates a simple yet effective approach to sampling from the space of learned representations. Naively, we assume that the set of latent codes extracted from a given face image is representative of the set of visual \textit{traits} present in the image. Under this assumption, we identify the \textit{common traits} between two coded representations $C_a$ and $C_b$ (for images $a$ and $b$ respectively as simply the intersection: $C_c = C_a \cap C_b$, and the \textit{exclusive traits} as the codes in either set but not in both: $C_d = (C_a \setminus C_b) \cup (C_b \setminus C_a)$.

\begin{algorithm}[t]
\caption{Fast interpolation-based sampling}
\small
\begin{algorithmic}[1]
\Procedure{Interpolate}{\texttt{Ca[1..L], Cb[1..L]}}
    \State \texttt{Cc = set intersect of Ca and Cb}
    \State \texttt{Cd = symmetric set difference of Ca and Cb}
    \State \texttt{Cinterp = Cc}
    \For{\texttt{j = 1 to L-length(Cc)}}
        \State \texttt{c = uniform random element from Cd}
        \State \texttt{add c to Cinterp}
        \State \texttt{remove c from Cd}
    \EndFor
    \State \Return \texttt{Cinterp}
\EndProcedure
\label{alg:interpolation}
\end{algorithmic}
\end{algorithm}

Defining $C_c$ and $C_d$ in this way, it is possible to sample from large (but finite) space of equally valid interpolations between $a$ and $b$ by first initialising $C_{interp}:=C_c$, then iteratively sampling elements from $C_d$ (without replacement) and adding them to $C_{interp}$ until $|C_{interp}|=L$. At this point, $C_{interp}$ forms a new representation comprising all the features shared by $C_a$ and $C_b$ as well as a mixture of features which are not shared. $C_{interp}$ is finally passed to the permutation-invariant decoder to produce an interpolated image $x_{inter}$ .

It is also possible to interpolate (approximately) smoothly from one image to another using a similar process: Compute $C_c$ as before. Construct two arrays  (\textit{not} sets) $R_a$ and $R_b$ of equal length $|R|$ containing the elements in $C_a$ but not in $C_b$ and vice versa, respectively. Choose a random permutation of each array $R_a'$ and $R_b'$ respectively. Construct interpolated representation $C_{smooth}^t$ for parameter $0<t<|R|$ as the union of ($C_c$, the first $t$ elements of $R_a$, the final $|R|-t$ elements of $R-b$). This way, there exists a unique interpolated path through the latent space for each unique pair of permutations $R_a'$ and $R_b'$.

\subsection{Training}
We follow largely the same training procedure as \cite{unleashing}. The model is trained to jointly minimise the L1 reconstruction loss, LPIPS perceptual loss \cite{lpips} (a learned perceptual distance function based on a pre-trained CNN classifier), and a learned discriminator loss \cite{gan}. The discriminator is trained alongside the decoder to distinguish reconstructed images from real images. We employ differentiable augmentations (DiffAug) \cite{diffaug} and adaptive weight limiting \cite{adaptive-weight}, since they have been shown to further boost reconstruction and sample quality in generative adversarial models.

\subsection{Delayed Codebook Initialization}
In practice, applying vector-quantization from the beginning of the training run leads to a catastrophic failure in learning, with the reconstructions remaining unrecognisable even at the end of training. To solve this problem, we delay the application of the vector quantization step until training is already partially underway. This follows earlier recommendations for the robust training of VQ-VAE models \cite{vq-robust}. 

At a pre-determined training iteration $T_q$, we apply data-dependent codebook initialization using KMeans++ \cite{kmpp}. The cluster centroids are fit to a history of encoder outputs over the most recent $W$ training iterations, where $W$ is a hyper-parameter chosen to trade-off time complexity against coverage of the latent space. In practice (Section \ref{sec:experiments}), we re-initialize the codebook more than once over successive windows of length $W$, finding that repeating codebook initialization in this way leads to better codebook usage.

\subsection{Probing the Learned Representations}

After training a PI-VQ model to auto-encode images, we wish to probe the learned representations for interpretable or semantic content. To achieve this, we train a binary logistic regression model for each of 15 ground-truth binary attributes (Fig. \ref{fig:logistic-regression} of FFHQ. Where the raw annotations are not binary (e.g. age), binary attributes are derived from raw annotations, e.g. "age\_senior". We treat the presence or absence of each latent code as a separate input variable (taking on the value 0 or 1 respectively), totalling one input variable for each code in the codebook. As a baseline predictor, we predict the most frequent label for each attribute (e.g. ``not has\_beard''). Section \ref{sec:experiments} discusses the findings of our probing experiment.

\section{Results and Discussion}
\label{sec:experiments}
\subsection{Setup}
\textbf{Datasets:} We repeat all our training runs for three face datasets : CelebA 64x64 \cite{celeba}, CelebaA-HQ 256x256 \cite{karras2017progressive} and FFHQ 256x256 \cite{ffhq}. We evaluate sample quality on the three datasets in terms of FID \cite{fid}, precision, recall \cite{precision-recall}, density and coverage \cite{density-coverage}. We compare against results reported in the literature where available. We note that some scores for density and coverage are missing in our comparisons with the state-of-the-art in cases where results were not reported, and we are unable to reproduce the results ourselves with our available computational resources.

\textbf{Architecture:} In the autoencoder, we use the same convolutional down-sampling and up-sampling architecture as \cite{unleashing} in the encoder and decoder respectively. We additionally include a 4-layer transformer encoder following the down-sampling and preceding the up-sampling modules, as illustrated in Fig.~\ref{fig:architecture}. The transformer embedding dimension is 256 and the FFN hidden dimension is 1024 (128 and 512 respectively for CelebA 64x64). We use the same discriminator and differentiable augmentations as \cite{unleashing}, since these are shown to improve visual quality.

\textbf{Hyperparameters:}  The transformer layers have a feedforward dimension equal to 4 times the input dimension. We use a relatively small batch size of 4 (increasing to 16 for CelebA 64x64) with exponential model averaging with $beta=0.995$ following \cite{unleashing}. We choose code length $L=512$ and codebook size $K=4096$ such that the information bottleneck is comparable to a conventional VQ-GAN with similar architecture (such as the one used in \cite{unleashing}): a conventional discrete bottleneck with $L=256$ latent codes and codebook size $K=1024$ corresponds to $256\log_2 1024=2560$ bits of information. A permutation-invariant discrete bottleneck with $L=512$ and $K=4096$ achieves a comparable bottleneck of approximately $2221$ bits, computed from our derived upper bound (Equation \ref{eq:matching-capacity}, Figure \ref{fig:capacity}). We reduce this to $L=32$ and $K=512$ for CelebA 64x64 due to the much smaller image size. Further training details are in supplemental.

\subsection{Codebook Usage and Bottleneck Capacity}

\begin{table}[t]
    \centering
    \scalebox{0.85}{
    \begin{tabular}{lcc}
        \toprule
        Quantization method & $K_{\mathrm{img}}$ (max) $\uparrow$ & Capacity (bits) $\uparrow$ \\
        \midrule
        Nearest & 49 & 614 \\
        Matching (ours) & \textbf{512} & \textbf{2221} \\
        \bottomrule
    \end{tabular}
    }
    \caption{Upper bound of latent bottleneck information capacity: comparison between nearest-neighbour and matching quantization for models trained on CelebA-HQ. $K_{\mathrm{img}}$ is included as the maximum per-image codebook usage observed over the entire dataset. Capacity is rounded to the nearest bit.}
    \label{tab:bottleneck-usage}
\end{table}

In order to verify empirically that our proposed matching quantization technique increases effective information capacity, we train a model on CelebA with identical settings, with the exception that the ``nearest neighbour'' vector quantization method is used in place of matching quantization. We find that the maximum per-image codebook usage $K_{\mathrm{img}}$ over the entire dataset is only $49$ out of the maximum $512$ when the ``nearest'' approach is used, indicating a large number of repeated codes per image. In comparison, the proposed ``matching'' technique achieves maximal per-image codebook usage by design (equal to $L$, which is 512 for our experiments), resulting in over $3.5\times$ the effective upper limit on information capacity (Table \ref{tab:bottleneck-usage}).

\subsection{Sample Quality and Diversity}

\begin{table}[t]
    \centering
    \scalebox{0.8}{
      \begin{tabular}{lccccc}
        \toprule
        Method & FID$\downarrow$ & P$\uparrow$ & R$\uparrow$ & D$\uparrow$ & C $\uparrow$ \\
        \midrule
         FFHQ 256x256 & & & & & \\
        \quad StyleGAN-XL \cite{sauer2022stylegan} & \textbf{2.2} & 0.80 & 0.39 & 0.86 & 0.73 \\
        \quad StyleGAN2 \cite{9156570} & 3.8 & 0.69 & 0.40 & 1.12 & \textbf{0.80} \\
        \quad VQ-GAN (UT) \cite{unleashing} & 7.1 & 0.69 & \textbf{0.48} & 1.06 & 0.77 \\
        \quad VQ-GAN (TT) \cite{taming} & 9.6 & 0.64 & 0.29 & 0.89 & 0.59 \\
        \quad VDVAE \cite{child2020very}& 28.5 & 0.59 & 0.20 & 0 .80 & 0.50 \\
        \quad PR-BigGAN (R) \cite{verine2023precision} & 35.2 & 0.78 & 0.10 & 0.89 & 0.60 \\
        \quad PR-BigGAN (P) \cite{verine2023precision} & 38.2 & \textbf{0.84} & 0.08 & \textbf{1.15} & 0.63 \\
        \quad BigGAN \cite{donahue2019large} & 41.4 & 0.66 & 0.10 & 0.52 & 0.47 \\
        \midrule
        \quad Ours & 23.4 & 0.69 & 0.19 & 1.02 & 0.55 \\
        \midrule
        CelebA 64x64 & & & & & \\
        \quad ADM-IP \cite{ning2023input} & \textbf{1.5} & 0.23 & \textbf{0.65} & 0.88 & 0.24 \\
        \quad PR-BigGAN (R) \cite{verine2023precision}  & 6.0 & 0.78 & 0.56 & 0.88 & \textbf{0.50} \\
        \quad BigGAN \cite{donahue2019large} & 9.2 & 0.78 & 0.51 & 0.89 & 0.48 \\
        \quad PR-BigGAN (P)\cite{verine2023precision}  & 22.5 & \textbf{0.84} & 0.26 & \textbf{1.21} & 0.43 \\
        \midrule
        \quad Ours  & 73.2 & 0.58 & 0.23 & 0.60 & 0.44 \\
        \midrule
        CelebA-HQ 256x256 & & & & & \\
        \quad RDM \cite{teng2023relay} & \textbf{3.2} & 0.77 & \textbf{0.55} & - & - \\
        \quad LDM-4 \cite{Rombach_2022_CVPR} & 5.1 & 0.72 & 0.49 & - & - \\
        \quad WaveDiff \cite{Phung_2023_CVPR} & 5.9 & -  & 0.37 & - & - \\
        \quad DDGAN \cite{xiao2021tackling} & 7.6 & - & 0.36 & - & - \\
        \midrule
        \quad Ours  &  22.8 &  \textbf{0.85} & 0.10 & 1.88 & 0.69 \\
        \midrule
        Average (SOTA)  &  2.3 & 0.82 & 0.56 & 1.18* & 0.65* \\
        Average (Ours) &  39.8 & 0.71 & 0.17 & 0.81* & 0.50* \\
        \bottomrule
    \end{tabular}}
    \caption{Comparison with state-of-the-art in FID, precision (P), recall (R), density (D), and coverage (C) scores for image quality. Best result for each dataset and metric is in bold. *Averages are taken over only the results where both SOTA and our results are available for comparison.}
    \label{tab:sample-quality}
\end{table}

To investigate whether the learned permutation-invariant representations support coherent image synthesis, we evaluate interpolation-based sampling using 5 metrics across 50,000 sampled images \footnote{In the case of CelebA-HQ, we only sample 30,000 images since the reference dataset has only 30,000 ground-truth images with which to compare.} synthesised from each of our 3 models trained with matching quantization: FID \cite{fid}, precision, recall \cite{precision-recall}, density, and coverage \cite{density-coverage}. Where available, we compare these values against state-of-the-art results reported in the literature. We choose to evaluate along a broad range of metrics in order to facilitate better comparisons deeper discussion.

We observe that PI-VQ approaches competitive scores in precision, density and coverage, achieving neither the best nor the worst result for each dataset across the majority of metrics. A more in-depth interpretation of these scores is given below:

\begin{enumerate}
    \item The high density scores on CelebA-HQ and FFHQ indicate that the model is good at sampling from regions of the dataset which are densely packed \cite{density-coverage}. It should be noted that $\mathbb{E}(density)=1$ if the synthetic distribution is identical to the real distribution \cite{density-coverage}, so our model achieves a \textit{higher sampler density than the underlying data} on the 256x256 resolution datasets.
    \item Coverage measures the fraction of real samples whose neighbourhoods contain at least one synthetic sample \cite{density-coverage}. The high coverage in the case of our method indicates that the model has good coverage of the data. 
    \item The lower FID and recall scores suggest that the permutation-invariant bottleneck may limit the representation's capacity to capture the full diversity of the data distribution. We discuss this trade-off in Section \ref{sec:limitations}.
\end{enumerate}




\subsection{Qualitative Results}

In this subsection, we examine the structure of the learned representation space by visualising interpolations between encoded images. We also demonstrate that our approach, while being fundamentally discrete in nature, can approximate a smooth interpolation between two images.

Figure \ref{fig:smooth-interpolation} shows 2 valid interpolations between two pairs of images (for a total of 4 interpolations) sampled using our smooth interpolation approach. The model successfully captures multiple plausible paths through the representation space, demonstrating that while individual codes are discrete, the combinatorial structure of the latent space permits approximate smoothness. This suggests the learned codes capture separable, compositional visual features. The number of possible interpolation paths for a given image pair is equal to $(|R|!)^2$, which likely contributes to the ability to sample densely from the 256x256 datasets.

\begin{figure}[t]
    \centering
    \includegraphics[width=0.95\linewidth]{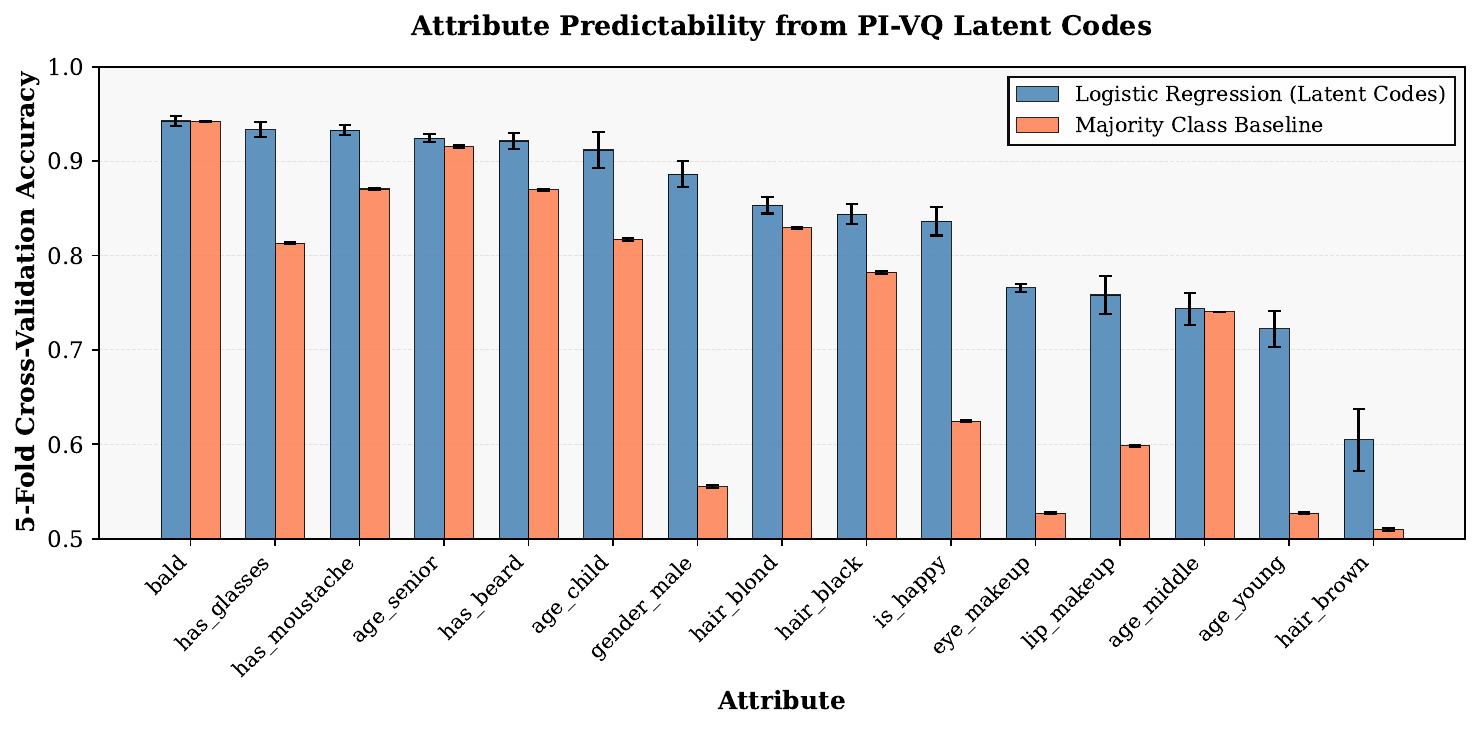}
    \caption{Logistic regression accuracy with error bars (5-way cross-validated) for predicting ground-truth FFHQ annotations based on learned permutation-invariant representations. For each attribute, we compare against the baseline accuracy (always predicting the majority class).}
    \label{fig:logistic-regression}
\end{figure}

To further probe the semantic content of the learned codebook entries, we fit logistic regression models to predict the ground-truth FFHQ annotations, treating the presence or absence of each unique latent code as a dummy binary variable (for a total number of variables equal to the codebook size). Figure \ref{fig:logistic-regression} indicates the 5-fold cross-validation accuracy of each regression model for the top 15 most ``predictable'' attributes, side-by-side with the majority class predictor (e.g. always predicting ``not bald'' as the most frequent annotation for hair presentation in FFHQ).

We find that, for certain attributes such as ``gender\_male'', ``is\_happy'' and ``eye\_makeup'', the logistic regression is significantly more accurate than the baseline. This indicates that the semantic codes capture separable, interpretable features which are predictive of human-annotated attributes, both global (e.g. gender) and local (e.g. eye\_makeup). Meanwhile, for some attributes, the logistic regression model is not much better than predicting the majority class (e.g. ``bald'', ``age\_middle'', ``age\_senior''), indicating that human annotations aren't universally captured by the learned codes in a linearly separable way.

\section{Limitations}
\label{sec:limitations}
The desirable properties demonstrated in Section~\ref{sec:experiments} (direct interpolation, $\mathcal{O}(1)$-time sampling, global codes) follow directly from the intended constraints of the proposed architecture. However, the design of our permutation-invariant method comes with a number of limitations, and our qualitative results reflect this trade-off.

We observe that while precision, density and coverage scores are comparable to the state-of-the-art, our method falls behind on FID and recall. The low recall (indicating mode-dropping) is likely a consequence of two factors. The permutation-invariant bottleneck has fundamentally lower information capacity than position-dependent alternatives, limiting the representation's ability to capture rare or fine-grained features. While the proposed matching quantization technique is effective in improving the effective information content of latent codes, it does not entirely prevent codebook collapse. Future work could combine our approach with techniques targeting codebook collapse directly.

Finally, the permutation-invariance constraint is inherently suited to global features and may struggle with fine-grained spatial details, which could contribute to perceptual differences captured by FID. Our evaluation focuses on aligned face datasets, where spatial structure is fixed. The approach is best understood as complementary to standard VQ methods—appropriate when global, interpretable discrete features are prioritised over spatial fidelity.
\section{Conclusion}
We have investigated the consequences of removing positional information from discrete image representations. The proposed PI-VQ architecture enforces permutation-invariance on learned codes, and this constraint encourages codes to capture global, semantic features rather than position-dependent patterns. As a consequence, the learned representations support direct interpolation between images and $\mathcal{O}(1)$ sampling without requiring a learned prior.

We introduced matching quantization as an alternative to nearest-neighbour vector quantization suited to the permutation-invariant setting, increasing the effective information capacity of the bottleneck by a factor of $3.5\times$. Our experiments on aligned face datasets demonstrate competitive precision, density and coverage metrics, while the worse recall and FID point to inherent trade-offs in position-free representations.

Future work could explore improved codebook utilisation, extension to higher resolutions, and application to other spatially-aligned domains beyond faces. Permutation-invariant discrete representation learning setting may also prove suitable for other modalities with global, compositional features. Our work aims to provide a useful foundation for further exploration of position-free discrete representation learning.

%
%
%

\bibliographystyle{splncs04}
\bibliography{main}
\end{document}


\setcounter{figure}{5}
\setcounter{table}{3}
\setcounter{equation}{4}
\setcounter{section}{6}
\setcounter{page}{9}
\maketitle

\section{Derivation of Information Capacities}

\subsection{Nearest-Neighbour VQ with Limited per-Image Codebook Usage}
In our experiments, we observed that applying naive nearest-neighbour vector-quantization \cite{vqvae} alongside PI-VQ results in a severely diminished per-image codebook usage $K_{img}$, the maximum of which over the whole training set was $49$ (where the technical maximum is $512$) when training on CelebA-HQ. This results in many repeated codes (since each image representation is of total length $512$), and hence redundancy in the discrete latent representation, which severely restricts the capacity of PI-VQ information bottleneck. 

Here we derive a precise upper limit on information capacity when using nearest-neighbor VQ, with an overall codebook usage (over the entire dataset) of $K_{data}$, representation length $L$, and maximum per-image codebook usage $K_{img}$. We use the same notation as in the main text.

In the best (maximum information) case, when representing an input image chosen at random from the dataset, the model selects $K_{img}$ codebook elements at random, without replacement, from the $K_{data}$ codebook elements. The space  $\mathcal{S}$  of possible values that this ``working subset'' can take on is of cardinality:

\begin{equation}
    |\mathcal{S}| = \begin{pmatrix*}K_{data}\\K_{img}\\ \end{pmatrix*}
\end{equation}

This is equivalent to stating that there are ``$K_{data}$ choose $K_{img}$'' ways to choose $K_{img}$ unique elements from a set of $K_{data}$ unique elements. 

Next, given a fixed working subset of length $K_{img}$, we determine the cardinality of the space $\mathcal{R}$ of representations of length $L$ consisting of only $K_{img}$ unique elements as:

\begin{equation}
    |\mathcal{R}| = \begin{pmatrix*}L+K_{img}-1\\K_{img}-1\\ \end{pmatrix*}
\end{equation}

This is equivalent to the number of ways one can distribute $L$ ``bars'' among $K_{img}$ ``stars'' using the combinatoric ``stars-and-bars'' graphical aid \cite{combinatorics}.

Finally, we give an upper bound on the cardinality of the space $\mathcal{M}$ of all possible representations as:

\begin{equation}
    |\mathcal{M}| < |\mathcal{S}|\times|\mathcal{R}|=\begin{pmatrix*}K_{data}\\K_{img}\\ \end{pmatrix*}\times\begin{pmatrix*}L+K_{img}-1\\K_{img}-1\\ \end{pmatrix*}
\end{equation}

We note that $|M|$ is strictly smaller than the product of the two because there exist instances of overlap between representations from two distinct ``working sets'', e.g. two distinct working sets both containing codebook entry $c$ can both represent an image as $c$ repeated $L$ times. This inequality is not an issue for our derivation since we are deriving an upper bound.

The upper bound on information capacity, in bits, is then:

\begin{align}
    \text{Capacity} &= \log_2\left[|\mathcal{S}|\times|\mathcal{R}|\right] \\
    &= \log_2\left[ \begin{pmatrix*}K_{data}\\K_{img}\\ \end{pmatrix*}\times\begin{pmatrix*}L+K_{img}-1\\K_{img}-1\\ \end{pmatrix*}\right]
    \label{eq:nn-capacity}
\end{align}

\subsection{Improved Capacity using Matching Quantization}

The proposed matching quantization approach ensures that representations of length $L$ always use exactly $L$ unique elements from the overall codebook. The space $\mathcal{M}$ of possible representations under this approach is then of cardinality:

\begin{equation}
    |\mathcal{M}| = \begin{pmatrix*}K_{data}\\L\\ \end{pmatrix*}
\end{equation}

This is because representing a given image is equivalent to choosing $L$ elements from a set of size $K_{data}$, to achieve which there are exactly ``$K_{data}$ choose $L$'' possible ways. We note this is equivalent to the naive quantization case with $K_{img}$ fixed at $L$. Thus the associated information capacity in bits is:

\begin{align}
    \text{Capacity} &= \log_2|\mathcal{M}| \\
     &= \log_2\left[\begin{pmatrix*}K_{data}\\L\\ \end{pmatrix*}\right]
\end{align}

\section{Training Details}

We train the PI-VQ run for 1,400,000 iterations for the 256x256 datasets and 400,000 iterations for CelebA 64x64. Following earlier work with VQ-GAN \cite{taming,unleashing}, we begin adversarial training part-way though the training run (iteration $500,000$ onward, or $100,000$ onward for CelebA 64x64).

\textbf{Codebook initialization and reset:} We set $T_q=60,000$ for FFHQ and CelebA-HQ. Due to the smaller overall number of training iterations on CelebA 64x64, we set $T_q=25,000$. In each case, $T_q$ is chosen to take place before the onset of adversarial training, and after at least one complete iteration through the entire training set.
\bibliographystyle{IEEEtran}
\bibliography{main.bib}